# A Study on Fuzzy Systems

**Michael Gr. Voskoglou**

School of Technological Applications, Graduate Technological Educational Institute  (T. E. I.), Patras 263 34, Greece
E-mail: mvosk@hol.gr,  Home page URL:  http://eclass.teipat.gr/eclass/courses/523102

**Abstract**   In the present paper we use principles of fuzzy logic to develop a general model representing several processes in a system's operation characterized by a degree of vagueness and/or uncertainty. For this, the main stages of the corresponding process are represented as fuzzy subsets of a set of linguistic labels characterizing the system's performance at each stage. We also introduce three alternative measures of a fuzzy system's effectiveness connected to our general model. These measures include the system's total possibilistic uncertainty, the Shannon's entropy properly modified for use in a fuzzy environment and the "centroid" method in which the coordinates of the center of mass of the graph of the membership function involved provide an alternative measure of the system's performance. The advantages and disadvantages of the above measures are discussed and a combined use of them is suggested for achieving a worthy of credit mathematical analysis of the corresponding situation. An application is also developed for the Mathematical Modelling process illustrating the use of our results in practice.

## 1. Introduction

A *system* is a set of interacting or interdependent components forming an integrated whole. A system comprises multiple views such as planning, analysis, design, implementation, deployment, structure, behavior, input and output data, etc. As an interdisciplinary and multi-perspective domain systems' theory brings together principles and concepts from ontology, philosophy of science, information and computer science, mathematics, as well as physics, biology, engineering, social and cognitive sciences, management and economics, strategic thinking, fuzziness and uncertainty, etc. Thus, it serves as a bridge for an interdisciplinary dialogue between autonomous areas of study. The emphasis with systems' theory shifts from parts to the organization of parts, recognizing that interactions of the parts are not static and constant, but dynamic processes. Most systems share common characteristics including structure, behaviour, interconnectivity (the various parts of a system have functional and structural relations to each other), sets of functions, etc. We scope a system by defining its boundary; this means choosing which entities are inside the system and which are outside, part of the environment.

The *systems' modelling* is a basic principle in engineering, in natural and in social sciences. When we face a problem concerning a system's operation (e.g. maximizing the productivity of an organization, minimizing the functional costs of a company, etc) a model is required to describe and represent the system's multiple views. The model is a simplified representation of the basic characteristics of the real system including only its entities and features under concern. In this sense, no model of a complex system could include all features and/or all entities belonging to the system.

In fact, in this way the model's structure could become very complicated and therefore its use in practice could be very difficult and sometimes impossible. Therefore the construction of the model usually involves a deep abstracting process on identifying the system's dominant variables and the relationships governing them. The resulting structure of this action is known as the *assumed real system* (see Figure 1). The model, being an abstraction of the assumed real system, identifies and simplifies the relationships among these variables in a form amenable to analysis.

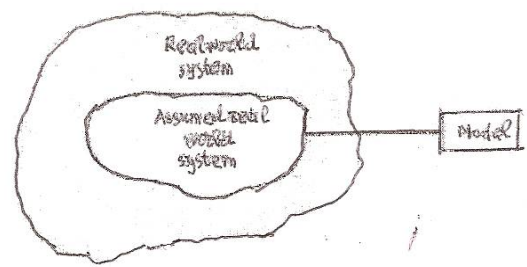

**Figure 1.**  A graphical representation of the modelling process

A system can be viewed as a bounded transformation, i.e. as a process or a collection of processes that transforms inputs into outputs with the very broad meaning of the concept. For example, an output of a passengers' bus is the movement of people from departure to destination.

Many of these processes are frequently characterized by a degree of vagueness and/or uncertainty. For example, during the processes of learning, of reasoning, of problem-solving, of modelling, etc, the human cognition utilizes in general concepts that are inherently graded and therefore fuzzy. On the other hand, from the teacher's point of view there usually exists an uncertainty about the degree of students' success in each of the stages of the corresponding didactic situation.

There used to be a tradition in science and engineering of turning to probability theory when one is faced with a problem in which uncertainty plays a significant role. This transition was justified when there were no alternative tools for dealing with the uncertainty. Today this is no longer the case. *Fuzzy logic*, which is based on fuzzy sets theory introduced by Zadeh[17] in 1965, provides a rich and meaningful addition to standard logic. The applications which may be generated from or adapted to fuzzy logic are wide-ranging and provide the opportunity for modelling under conditions which are inherently imprecisely defined, despite the concerns of classical logicians. Many systems may be modelled, simulated and even replicated with the help of fuzzy logic, not the least of which is human reasoning itself (e.g.[3],[4],[7],[8],[12],[14],[15],[16] etc)

A real test of the effectiveness of an approach to uncertainty is the capability to solve problems which involve different facets of uncertainty. Fuzzy logic has a much higher problem solving capability than standard probability theory. Most importantly, it opens the door to construction of mathematical solutions of computational problems which are stated in a natural language. In contrast, standard probability theory does not have this capability, a fact which is one of its principal limitations.

All these gave us the impulsion to introduce principles of fuzzy logic to describe in a more effective way a system's operation in situations characterized by a degree of vagueness and/or uncertainty.

For general facts on fuzzy sets and on uncertainty theory we refer freely to the book of Klir and Folger[1].

## 2. The General Fuzzy Model

Assume that one wants to study the behavior of a system's $n$ entities (objects), $n \geq 2$, during a process involving vagueness and/or uncertainty. Denote by $S_i$, $i=1,2,3$ the main stages of this process and by $a, b, c, d,$ and $e$ the linguistic labels of very low, low, intermediate, high and very high success respectively of a system's entity in each of the $S_i$'s. Set

$$U = \{a, b, c, d, e\}.$$

We are going to attach to each stage $S_i$ a fuzzy subset, $A_i$ of $U$. For this, if $n_{ia}$, $n_{ib}$, $n_{ic}$, $n_{id}$ and $n_{ie}$ denote the number of entities that faced very low, low, intermediate, high and very high success at stage $S_i$ respectively, $i=1,2,3$, we define the *membership function* $m_{Ai}$ for each $x$ in $U$, as follows:

$$m_{A_i}(x) = \begin{cases} 1, & \text{if } \frac{4n}{5} < n_{ix} \leq n \\ 0{,}75, & \text{if } \frac{3n}{5} < n_{ix} \leq \frac{4n}{5} \\ 0{,}5, & \text{if } \frac{2n}{5} < n_{ix} \leq \frac{3n}{5} \\ 0{,}25, & \text{if } \frac{n}{5} < n_{ix} \leq \frac{2n}{5} \\ 0, & \text{if } 0 \leq n_{ix} \leq \frac{n}{5} \end{cases}$$

Then the fuzzy subset $A_i$ of $U$ corresponding to $S_i$ has the form:

$$A_i = \{(x, m_{Ai}(x)): x \in U\}, i=1, 2, 3.$$

In order to represent all possible *profiles (overall states)* of the system's entities during the corresponding process we consider a *fuzzy relation*, say $R$, in $U^3$ of the form:

$$R = \{(s, m_R(s)): s=(x, y, z) \in U^3\}.$$

We assume that the stages of the process that we study are depended to each other. This means that the degree of system's entity success in a certain stage depends upon the degree of its success in the previous stages, as it usually happens in practice. Under this hypothesis and in order to determine properly the membership function $m_R$ we give the following definition:

*Definition*: A profile $s=(x, y, z)$, with $x, y, z$ in $U$, is said to be *well ordered* if $x$ corresponds to a degree of success equal or greater than $y$ and $y$ corresponds to a degree of success equal or greater than $z$.

For example, *(c, c, a)* is a well ordered profile, while *(b, a, c)* is not.

We define now the *membership degree* of a profile $s$ to be

$$m_R(s) = m_{A_1}(x) m_{A_2}(y) m_{A_3}(z)$$

if $s$ is well ordered, and $0$ otherwise.

In fact, if for example the profile *(b, a, c)* possessed a nonzero membership degree, how it could be possible for an object that has failed during the middle stage, to perform satisfactorily at the next stage?

Next, for reasons of brevity, we shall write $m_s$ instead of $m_R(s)$.

Then the *probability* $p_s$ of the profile s is defined in a way analogous to crisp data, i.e. by

$$P_s = \frac{m_s}{\sum_{s \in U^3} m_s}.$$

We define also the *possibility* $r_s$ of $s$ by

$$r_s = \frac{m_s}{\max\{m_s\}},$$

where $\max\{m_s\}$ denotes the maximal value of $m_s$, for all $s$ in $U^3$. In other words the possibility of $s$ expresses the "relative membership degree" of $s$ with respect to $\max\{m_s\}$.

Assume further that one wants to study the *combined results* of behaviour of $k$ different groups of a system's entities, $k \geq 2$, during the same process.

For this we introduce the *fuzzy variables* $A_1(t)$, $A_2(t)$ and $A_3(t)$ with $t=1, 2,..., k$. The values of these variables represent fuzzy subsets of $U$ corresponding to the stages of the process for each of the $k$ groups; e.g. $A_1(2)$ represents the fuzzy subset of $U$ corresponding to the first stage of the process for the second group ($t=2$). It becomes evident that, in order to measure the degree of evidence of the combined results of the $k$ groups, it is necessary to define the probability $p(s)$ and the possibility $r(s)$ of each profile $s$ with respect to the membership degrees of $s$ for all groups. For this reason we introduce the *pseudo-frequencies*

$$f(s) = \sum_{t=1}^{k} m_s(t)$$

and we define the probability of a profile s by

$$p(s) = \frac{f(s)}{\sum_{s \in U^3} f(s)}.$$

We also define the possibility of s by

$$r(s) = \frac{f(s)}{\max\{f(s)\}},$$

where $\max\{f(s)\}$ denotes the maximal pseudo-frequency.

Obviously the same method could be applied when one wants to study the combined results of behaviour of a group during $k$ different situations.

## 3. Fuzzy Measures of a System's Effectiveness

There are *natural* and *human-designed* systems. Natural systems may not have an apparent objective, but their outputs can be interpreted as purposes. On the contrary, human-designed systems are made with purposes that are achieved by the delivery of outputs. Their parts must be related, i.e. they must be designed to work as a coherent entity.

The most important part of a human-designed system's study is probably the assessment, through the model representing it, of its performance. In fact, this could help the system's designer to make all the necessary modifications/improvements to the system's structure in order to increase its effectiveness.

In this article we'll present three fuzzy measures of a system's effectiveness connected to the general fuzzy model developed above. The advantages and disadvantages of these measures will be also discussed and an application for the problem solving process will be presented illustrating our results.

The amount of information obtained by an action can be measured by the reduction of uncertainty resulting from this action. Accordingly a system's uncertainty is connected to its capacity in obtaining relevant information. Therefore a measure of uncertainty could be adopted as a measure of a system's effectiveness in solving related problems.

Within the domain of possibility theory uncertainty consists of *strife (or discord)*, which expresses conflicts among the various sets of alternatives, and *non-specificity (or imprecision)*, which indicates that some alternatives are left unspecified, i.e. it expresses conflicts among the sizes (cardinalities) of the various sets of alternatives ([2]; p.28).

Strife is measured by the function $ST(r)$ on the ordered possibility distribution

$$r: r_1=1 \geq r_2 \geq \ldots \geq r_n \geq r_{n+1}$$

of a group of a system's entities defined by

$$ST(r) = \frac{1}{\log 2}[\sum_{i=2}^{m}(r_i - r_{i+1})\log \frac{i}{\sum_{j=1}^{i} r_j}].$$

Non-specificity is measured by the function

$$N(r) = \frac{1}{\log 2}[\sum_{i=2}^{m}(r_i - r_{i+1})\log i].$$

The sum $T(r) = ST(r) + N(r)$ is a measure of the *total possibilistic uncertainty* for ordered possibility distributions. The lower is the value of $T(r)$, which means greater reduction of the initially existing uncertainty, the better the system's performance.

Another fuzzy measure for assessing a system's performance is the well known from classical probability and information theory *Shannon's entropy*[6]. For use in a fuzzy environment, this measure is expressed in terms of the Dempster-Shafer mathematical theory of evidence in the form:

$$H = -\frac{1}{\ln n}\sum_{s=1}^{n} m_s \ln m_s$$

([2], p. 20).

In the above formula $n$ denotes the total number of the system's entities involved in the corresponding process. The sum is divided by $\ln n$ (the natural logarithm of $n$) in order to be normalized. Thus $H$ takes values in the real interval $[0, 1]$. The value of $H$ measures the system's total *probabilistic uncertainty* and the associated to it information. Similarly with the total possibilistic uncertainty, the lower is the final value of $H$, the better the system's performance.

An advantage of adopting $H$ as a measure instead of $T(r)$ is that $H$ is calculated directly from the membership degrees of all profiles $s$ without being necessary to calculate their probabilities $p_s$. In contrast, the calculation of $T(r)$ presupposes the calculation of the possibilities $r_s$ of all profiles first. However, according to Shackle[5] human reasoning can be formalized more adequately by possibility rather, than by probability theory. But, as we have seen in the previous section, the possibility is a kind of "relative probability". In other words, the "philosophy" of possibility is not exactly the same with that of probability theory. Therefore, on comparing the effectiveness of two or more systems by these two measures, one may find non compatible results in boundary cases, where the systems' performances are almost the same.

Another popular approach is the *"centroid" method*, in which the centre of mass of the graph of the membership function involved provides an alternative measure of the system's performance.

For this, given a fuzzy subset
$$A = \{(x, m(x)): x \in U\}$$
of the universal set *U* with membership function

*m: U → [0, 1]*, we correspond to each $x \in U$ an interval of values from a prefixed numerical distribution, which actually means that we replace *U* with a set of real intervals. Then, we construct the graph *F* of the membership function *y=m(x)*.

There is a commonly used in fuzzy logic approach to measure performance with the pair of numbers $(x_c, y_c)$ as the coordinates of the *centre of mass*, say $F_c$, of the graph *F*, which we can calculate using the following well-known [10] formulas:

$$x_c = \frac{\iint_F x\,dx\,dy}{\iint_F dx\,dy}, \quad y_c = \frac{\iint_F y\,dx\,dy}{\iint_F dx\,dy} \qquad (1).$$

For example, assume that the set U of the linguistic labels defined in the previous section characterizes the performance of a group of students. When a student obtains a mark, say y, then his/her performance is characterized as very low (a) if $y \in [0, 1)$, as low (b) if $y \in [1, 2)$, as intermediate (c) if $y \in [2, 3)$, as high (d) if $y \in [3, 4)$ and as very high (e) if $y \in [4, 5]$ respectively. In this case the graph F of the corresponding fuzzy subset of U is the bar graph of Figure 2

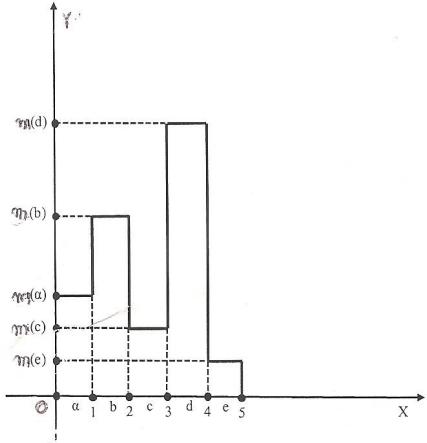

**Figure 2.** Bar graphical data representation

It is easy to check that, if the bar graph consists of *n* rectangles (in Figure 2 we have n=5), the formulas (1) can be reduced to the following formulas:

$$x_c = \frac{1}{2}\left(\frac{\sum_{i=1}^{n}(2i-1)y_i}{\sum_{i=1}^{n}y_i}\right), \quad y_c = \frac{1}{2}\left(\frac{\sum_{i=1}^{n}y_i^2}{\sum_{i=1}^{n}y_i}\right) \quad (2)$$

Indeed, in this case $\iint_F dx\,dy$ is the total mass of the system which is equal to $\sum_{i=1}^{n} y_i$, $\iint_F x\,dx\,dy$ is the moment about the y-axis which is equal to

$$\sum_{i=1}^{n}\iint_{F_i} x\,dx\,dy = \sum_{i=1}^{n}\int_0^{y_i} dy \int_{i-1}^{i} x\,dx = \sum_{i=1}^{n} y_i \int_{i-1}^{i} x\,dx = \frac{1}{2}\sum_{i=1}^{n}(2i-1)y_i,$$

and $\iint_F y\,dx\,dy$ is the moment about the y-axis which is equal to

$$\sum_{i=1}^{n}\iint_{F_i} y\,dx\,dy = \sum_{i=1}^{n}\int_0^{y_i} y\,dy \int_{i-1}^{i} dx =$$

$$\sum_{i=1}^{n}\iint_{F_i} x\,dx\,dy = \sum_{i=1}^{n}\int_0^{y_i} dy \int_{i-1}^{i} x\,dx = \sum_{i=1}^{n}\int_0^{y_i} y\,dy = \frac{1}{2}\sum_{i=1}^{n} y_i^2.$$

From the above argument, where $F_i$, i=1,2,…,n, denote the n rectangles of the bar graph, it becomes evident that the transition from (1) to (2) is obtained under the assumption that all the intervals have length equal to 1 and that the first of them is the interval[0, 1].

In our case (n=5) formulas (2) are transformed into the following form:

$$x_c = \frac{1}{2}\left(\frac{y_1 + 3y_2 + 5y_3 + 7y_4 + 9y_5}{y_1 + y_2 + y_3 + y_4 + y_5}\right),$$

$$y_c = \frac{1}{2}\left(\frac{y_1^2 + y_2^2 + y_3^2 + y_4^2 + y_5^2}{y_1 + y_2 + y_3 + y_4 + y_5}\right).$$

Normalizing our fuzzy data by dividing each m(x), $x \in U$, with the sum of all membership degrees we can assume without loss of the generality that
$$y_1+y_2+y_3+y_4+y_5 = 1.$$

Therefore we can write:

$$x_c = \frac{1}{2}(y_1 + 3y_2 + 5y_3 + 7y_4 + 9y_5),$$
$$y_c = \frac{1}{2}(y_1^2 + y_2^2 + y_3^2 + y_4^2 + y_5^2) \qquad (3)$$

with $y_i = \dfrac{m(x_i)}{\sum_{x \in U} m(x)}$, where $x_1$= a, $x_2$=b, $x_3$= c,

$x_4$ = d and $x_5$ = e.

But
$$0 \le (y_1-y_2)^2 = y_1^2+y_2^2-2y_1y_2,$$
therefore
$$y_1^2+y_2^2 \ge 2y_1y_2$$
with the equality holding if, and only if, $y_1=y_2$.

In the same way one finds that
$$y_1^2+y_3^2 \ge 2y_1y_3,$$
and so on. Hence it is easy to check that
$$(y_1+y_2+y_3+y_4+y_5)^2 \le 5(y_1^2+y_2^2+y_3^2+y_4^2+y_5^2),$$
with the equality holding if, and only if $y_1=y_2=y_3=y_4=y_5$.

But $y_1+y_2+y_3+y_4+y_5 =1$, therefore
$$1 \le 5(y_1^2+y_2^2+y_3^2+y_4^2+y_5^2) \quad (4),$$
with the equality holding if, and only if $y_1=y_2=y_3=y_4=y_5= \frac{1}{5}$.

Then the first of formulas (3) gives that $x_c = \frac{5}{2}$. Further, combining the inequality (4) with the second of formulas (3) one finds that
$$1 \le 10y_c, \text{ or } y_c \ge \frac{1}{10}.$$

Therefore the unique minimum for $y_c$ corresponds to the centre of mass $F_m(\frac{5}{2}, \frac{1}{10})$.

The ideal case is when $y_1=y_2=y_3=y_4=0$ and $y_5=1$. Then from formulas (3) we get that $x_c = \frac{9}{2}$ and $y_c = \frac{1}{2}$. Therefore the centre of mass in this case is the point $F_i(\frac{9}{2}, \frac{1}{2})$.

On the other hand the worst case is when $y_1=1$ and $y_2=y_3=y_4=y_5=0$. Then for formulas (3) we find that the centre of mass is the point $F_w(\frac{1}{2}, \frac{1}{2})$.

Therefore the "area" where the centre of mass $F_c$ lies is represented by the triangle $F_w F_m F_i$ of Figure 3

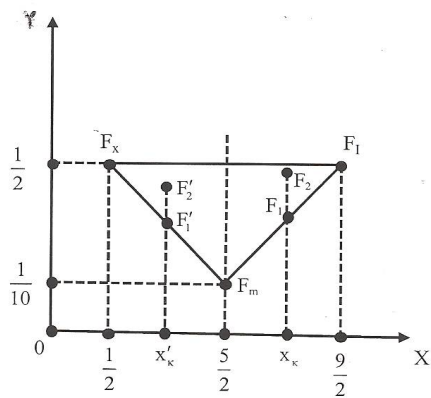

**Figure 3.** A graphical representation of the "area" of the centre of mass

Then from elementary geometric considerations it follows that for two groups of a system's objects with the same $x_c \geq 2.5$ the group having the centre of mass which is situated closer to $F_i$ is the group with the higher $y_c$; and for two groups with the same $x_c < 2.5$ the group having the centre of mass which is situated farther to $F_w$ is the group with the lower $y_c$.

Based on the above considerations it is logical to formulate our criterion for comparing the groups' performances in the following form:

• Among two or more groups the group with the biggest $x_c$ performs better.

• If two or more groups have the same $x_c \geq 2.5$, then the group with the higher $y_c$ performs better.

• If two or more groups have the same $x_c < 2.5$, then the group with the lower $y_c$ performs better.

From the above description it becomes clear that the application of the "centroid" method in practice is simple and evident and needs no complicated calculations in its final step. However, we must emphasize that this method treats differently the idea of a system's performance, than the two measures of uncertainty presented above do. In fact, the weighted average plays the main role in this method, i.e. the result of the system's performance close to its ideal performance has much more weight than the one close to the lower end. In other words, while the measures of uncertainty are dealing with the average system's performance, the "centroid" method is mostly looking at the quality of the performance. Consequently, some differences could appear in evaluating a system's performance by these different approaches. Therefore, it is argued that a combined use of all these (3 in total) measures could help the user in finding the ideal profile of the system's performance according to his/her personal criteria of goals.

## 4. Modelling the Process of Mathematical Modelling

In earlier papers we have developed models similar to the above general model for a more effective description of several situations involving fuzziness and/or uncertainty in the areas of Education (for the processes of Learning and of Problem Solving), of Artificial Intelligence (for Case-Based and Analogical Reasoning) and of Management (for evaluating the fuzzy data obtained by a market's research and for Decision Making); see for example[15] and its references. Notice also, that Subbotin et al., based on our fuzzy model for the process of Learning[12], have applied the "centroid" method on comparing students' mathematical learning abilities[7] and for measuring the scaffolding (assistance) effectiveness provided by the teacher to students[8], while Perdikaris ([3],[4]) has used the total possiblistic uncertainty and the Shannon's entropy for measuring student's geometrical reasoning skills in terms of the corresponding van Hieles' levels.

In this article we shall apply our general model developed above for the representation of the process of Mathematical Modelling (MM).

The representation of a system's operation through the use of a mathematical model is achieved by a set of mathematical relations (equalities, inequalities, etc) and functions properly related to each other.

It is well known (e.g.[9]; paragraph 1.4) that the stages of the MM process involve:

• *Analysis* of the given real world problem, i.e. understanding the statement and recognizing limitations, restrictions and requirements of the real system.

• *Mathematization*, i.e. *formulation* of the real situation in such a way that it will be ready for mathematical treatment (assumed real system, see first section) and *construction* of the model.

• *Solution* of the model, achieved by proper mathematical manipulation.

• *Validation* (control) of the model, usually achieved by reproducing through it the behavior of the real system under the conditions existing before the solution of the model (empirical results, special cases etc). A simulation model is also frequently used for this purpose as a secondary model.

• *Implementation* of the final mathematical results to the real system, i.e. "translation" of the mathematical solution obtained in terms of the corresponding real situation in order to reach the solution of the given real problem.

For the development of our fuzzy model for the MM process we consider a group of *n* modellers, n ≥ 2, working (each one individually) on the same modelling problem. In order to make our model technically simpler, we can, without loss of the generality, reduce the stages of the MM process to the following three:

$S_1$ : Analysis/Mathematization,
$S_2$ : Solution of the model,
$S_3$ : Validation/Implementation

In fact, the analysis of the given problem is an introductory stage of the MM process that can be naturally seen as being a sub step of mathematization. Further, the stage of implementation of the final mathematical results to the real system is an expected action following the validation of the model, which means that the joined stage of Validation/Implementation can be considered without loss of the generality as the final stage of the MM process.

To each of the $S_i$'s we attach a fuzzy subset, say $A_i$, of the set *U* of the linguistic labels considered in the second section defining also the membership function $m_{Ai}$ as we did in this section. The development of the rest of our model for the MM process relies then upon the lines of our general fuzzy model presented in detail in the two previous sections.

In order to illustrate the use of our results in practice, we performed the experiments presented in the next section.

## 5. Applications of the Model for MM

The following two experiments took place recently at the Graduate Technological Educational Institute (T. E. I.) of Patras in Greece. In the first of them our subjects were 35 students of the School of Technological Applications, i.e. future engineers, and our basic tool was a list of 10 problems (see Appendix) given to them for solution (time allowed 3 hours). Before starting the experiment we gave the proper instructions to students emphasizing among the others that we are interested for all their efforts (successful or not) during the MM process, and therefore they must keep records on their papers for all of them, at all stages of the MM process. This manipulation enabled as in obtaining realistic data from our experiment for each stage of the MM process and not only those based on students' final results that could be obtained in the usual way by graduating their papers.

Our characterizations of students' performance at each stage of the MM process involved:

• Negligible success, if they obtained (at the particular stage) positive results for less than 2 problems.
• Low success, if they obtained positive results for 2, 3, or 4 problems.
• Intermediate success, if they obtained positive results for 5, 6, or 7 problems.
• High success, if they obtained positive results for 8, or 9 problems.
• Very high success, if they obtained positive results for all problems.

Examining students' papers we found that 15, 12 and 8 students had intermediate, high and complete success respectively at stage $S_1$ of analysis/mathematization. Therefore we obtained that $n_{1a}=n_{1b}=0$, $n_{1c}=15$, $n_{1d}=12$ and $n_{1e}=8$. Thus, by the definition of the corresponding membership function given in the second section, $S_1$ is represented by a fuzzy subset of *U* of the form:

$A_1 = \{(a,0),(b,0),(c, 0,5),(d, 0,25),(e,0,.25)$,

In the same way we represented the stages $S_2$ and $S_3$ as fuzzy sets in *U* by

$A_2 = \{(a,0),(b,0),(c, 0,5),(d, 0,25),(e,0)\}$

and

$A_3 = \{(a, 0,25),(b, 0,25),(c, 0,25),(d,0),(e,0)\}$

respectively.

Next we calculated the membership degrees of the $5^3$ (ordered samples with replacement of 3 objects taken from 5) in total possible students' profiles as it is described in the second section (see column of $m_s(1)$ in Table 1). For example, for the profile

*s=(c, c, a)* one finds that
$m_s = m_{A_1}(c). m_{A_2}(c). m_{A_3}(a) = 0,5.0,5.0,25 = 0,06225$.

It is straightforward then to calculate in terms of the membership degrees the Shannon's entropy for the student group, which is $H \approx 0,289$.

Further, from the values of the column of $m_s(1)$ it turns out that the maximal membership degree of students' profiles is *0,06225*. Therefore the possibility of each *s* in $U^3$ is given by

$$r_s = \frac{m_s}{0,06225}.$$

Calculating the possibilities of all profiles (column of $r_s(1)$ in Table 1) one finds that the ordered possibility distribution for the student group is:

r: $r_1 = r_2 = 1$, $r_3 = r_4 = r_5 = r_6 = r_7 = r_8 = 0,5$, $r_9 = r_{10} = r_{11} = r_{12} = r_{13} = r_{14} = 0,258$, $r_{15} = r_{16} = \ldots = r_{125} = 0$.

Thus with the help of a calculator one finds that

$$ST(r) = \frac{1}{\log 2}[\sum_{i=2}^{14}(r_i - r_{i+1})\log \frac{i}{\sum_{j=1}^{i} r_j}]$$

$$\approx \frac{1}{0,301}[0,5\log\frac{2}{2} + 0,242\log\frac{8}{5} + 0,258\log\frac{14}{6,548}]$$

$$\approx 3,32 . 0,242 . 0,204 + 0,258 . 0,33 \approx 0,445. \text{ and}$$

$$N(r) = \frac{1}{\log 2}[\sum_{i=2}^{14}(r_i - r_{i+1})\log i]$$

$$= \frac{1}{\log 2}(0,5_{\log 2} + 0,242 \log 8 + 0,258 \log 14)$$

$$\approx 0,5 + 3,0,242 + 0,857.1,146 \approx 2,208.$$

Therefore we finally have that $T(r) \approx 2,653$

**Table 1.** Profiles with non zero membership degrees (The outcomes of the above Table were obtained with accuracy up to the third decimal point)

| $A_1$ | $A_2$ | $A_3$ | $m_s(1)$ | $r_s(1)$ | $m_s(2)$ | $r_s(2)$ | $f(s)$ | $r(s)$ |
|---|---|---|---|---|---|---|---|---|
| B | b | b | 0 | 0 | 0.016 | 0.258 | 0.016 | 0.129 |
| B | b | a | 0 | 0 | 0.016 | 0.258 | 0.016 | 0.129 |
| B | a | a | 0 | 0 | 0.016 | 0.258 | 0.016 | 0.129 |
| C | c | c | 0.062 | 1 | 0.062 | 1 | 0.124 | 1 |
| C | c | a | 0.062 | 1 | 0.062 | 1 | 0.124 | 1 |
| C | c | b | 0 | 0 | 0.031 | 0.5 | 0.031 | 0.25 |
| C | a | a | 0 | 0 | 0.031 | 0.5 | 0.031 | 0.25 |
| C | b | a | 0 | 0 | 0.031 | 0.5 | 0.031 | 0.25 |
| C | b | b | 0 | 0 | 0.031 | 0.5 | 0.031 | 0.25 |
| D | d | a | 0.016 | 0.258 | 0 | 0 | 0.016 | 0.129 |
| D | d | b | 0.016 | 0.258 | 0 | 0 | 0.016 | 0.129 |
| D | d | c | 0.016 | 0.258 | 0 | 0 | 0.016 | 0.129 |
| D | a | a | 0 | 0 | 0.016 | 0.258 | 0.016 | 0.129 |
| D | b | a | 0 | 0 | 0.016 | 0.258 | 0.016 | 0.129 |
| D | b | b | 0 | 0 | 0.016 | 0.258 | 0.016 | 0.129 |
| D | c | a | 0.031 | 0.5 | 0.031 | 0.5 | 0.062 | 0.5 |
| D | c | b | 0.031 | 0.5 | 0.031 | 0.5 | 0.062 | 0.5 |
| D | c | c | 0.031 | 0.5 | 0.031 | 0.5 | 0.062 | 0.5 |
| E | c | a | 0.031 | 0.5 | 0 | 0 | 0.031 | 0.25 |
| E | c | b | 0.031 | 0.5 | 0 | 0 | 0.031 | 0.25 |
| E | c | c | 0.031 | 0.5 | 0 | 0 | 0.031 | 0.25 |
| E | d | a | 0.016 | 0.258 | 0 | 0 | 0.016 | 0.129 |
| E | d | b | 0.016 | 0.258 | 0 | 0 | 0.016 | 0.129 |
| E | d | c | 0.016 | 0.258 | 0 | 0 | 0.016 | 0.129 |

A few days later we performed the same experiment with a group of 30 students of the School of Management and Economics. Working as above we found that

$A_1$={(a, 0),(b, 0,25),(c, 0,5),(d, 0 ,25),(e, 0)},
$A_2$={(a, 0,25),(b, 0,25),(c, 0,5),(d, 0),(e, 0)}
$A_3$={(a, 0,25),(b, 0,25),(c,0,25),(d, 0),(e, 0)}.

Then we calculated the membership degrees of all possible profiles of the student group (column of $m_s(2)$ in Table 1) and the Shannon's entropy, which is $H \approx 0,312$.

Since the maximal membership degree is again 0,06225, the possibility of each $s$ is given by the same formula as for the first group. Calculating the possibilities of all profiles (column of $r_s(2)$ in Table 1) one finds that the ordered possibility distribution of the second group is:

r: $r_1 = r_2 = 1$, $r_3 = r_4 = r_5 = r_6 = r_7 = r_8 = 0,5$, $r_9 = r_{10} = r_{11} = r_{12} = r_{13} = 0,258$, $r_{14} = r_{15} = \ldots = r_{125} = 0$

Finally, working in the same way as above one finds that $T(r) = 0,432+2,179 = 2,611$.

Therefore, since $2,611<2,653$, it turns out that the second group had in general a slightly better performance than the first one. Notice that the values of the Shannon's entropy lead to the opposite conclusion (since $0,312>0,289$), but this, as we have already explained in the third section, is not surprising in cases, where the difference between the performances of the two groups is very small. Further, using formulas (3) one can compare the performances of the two groups by the "centroid" method in each of the listed above stages of the MM process as follows:

Denote by $A_{ij}$ the fuzzy subset of U attached to the stage $S_j$, j=1,2,3, of the MM process with respect to the student group i, i=1,2.

At the first stage of analysis/mathematization we have
$A_{11}$ = {(a, 0),(b, 0),(c, 0,5),(d, 0,25),(e, 0,25)}
$A_{21}$= {(a, 0),(b, 0,25),(c, 0,5),(d , 0,25),(e, 0)}
and respectively

$$x_{c11} = \frac{1}{2}(5 \cdot 0,5+7 \cdot 0,25+9 \cdot 0,25) = 3,25$$

$$x_{c21} = \frac{1}{2}(3 \cdot 0,25+5 \cdot 0,5+7 \cdot 0,25) = 2,25.$$

By our criterion the first group demonstrates better performance.

At the second stage of solution we have:
$A_{12}$ = {(a, 0),(b, 0),(c, 0,5),(d, 0,25),(e, 0)},
$A_{22}$={(a, 0,25),(b, 0,25),(c, 0,5),(d, 0),(e, 0)}.

Normalizing the membership degrees in the first of the above fuzzy subsets of U (0,5 : 0,75 $\approx$ 0,67 and 0,25 : 0,75 $\approx$ 0,33) we get
$A_{12}$ = {(a, 0),(b, 0),(c, 0,67),(d, 0,33),(e, 0)},
$A_{22}$={(a, 0,25),(b, 0,25),(c, 0,5),(d, 0),(e, 0)}
and respectively

$$x_{c12} = \frac{1}{2}(5 \cdot 0,67+7 \cdot 0,33) = 5,66$$

$$x_{c22} = \frac{1}{2}(0,25+3 \cdot 0,25+5 \cdot 0,25) = 3,25.$$

By our criterion, the first group again demonstrates a significantly better performance.

Finally, at the third stage of validation/implementation we have
$A_{13}$= $A_{23}$ = {(a, 0,25),(b, 0,25),(c, 0,25),(d, 0),(e, 0)},
which obviously means that at this stage the performances of both groups are identical.

Based on our calculations we can conclude that the first group demonstrated a significantly better performance at the stages of analysis/mathematization and of solution, but performed identically with the second one at the stage of validation/implementation.

*Remark:* In earlier papers ([11],[13]) we have also developed a stochastic model for the representation of the MM process by applying a Markov chain on its stages. However, our stochastic model is self restricted to give quantitative information only for the MM process through the description of the *ideal behavior* of a group of modelers (i.e. how they must act for the solution of a problem and not how they really act in practice). In contrast, the above developed fuzzy model has the advantage of giving, apart of quantitative information, a qualitative/realistic view of the MM process through the calculation of the probabilities and/or possibilities of all possible modellers' profiles.

Nevertheless, the characterization of the modellers' performance in terms of a set of linguistic labels, which are fuzzy themselves, is a disadvantage of the fuzzy model, because this characterization depends on the user's personal criteria. A "live" example about this is the different evaluations for the two groups of modellers obtained in our classroom experiments by using our fuzzy measures for the MM skills. Therefore the stochastic could be used as a tool for the validation of the fuzzy model in an effort of achieving a worthy of credit mathematical analysis of the MM process.

## 6. Conclusions

The following conclusions can be drawn from the discussion performed in this paper:

- We developed a general fuzzy model for representing several processes in a system's operation involving vagueness and/or uncertainty.
- We presented 3 alternative methods of measuring a system's effectiveness connected to the above model.
- We applied our general fuzzy model for the description of the MM process. Our corresponding stochastic model developed in earlier papers could be used as a tool for the validation of the fuzzy model in achieving a worthy of credit mathematical analysis of the MM process.

## Appendix

List of the problems given for solution to students in our classroom experiments

Problem 1: We want to construct a channel to run water by folding the two edges of an orthogonal metallic leaf having sides of length 20cm and 32 cm, in such a way that they will be perpendicular to the other parts of the leaf. Assuming that the flow of the water is constant, how we can run the maximum possible quantity of the water?

Remark: The correct solution is obtained by folding the edges of the longer side of the leaf. Some students solved the problem by folding the edges of the other side and failed to realize (validation of the model) that their solution was wrong.

Problem 2: A car dealer has a mean annual demand of 250 cars, while he receives 30 new cars per month. The annual cost of storing a car is 100 euros and each time he makes a new order he pays an extra amount of 2200 euros for general expenses (transportation, insurance etc). The first cars of a new order arrive at the time when the last car of the previous order has been sold. How many cars must he order in order to achieve the minimum total cost?

Problem 3: An importation company codes the messages for the arrivals of its orders in terms of characters consisting of a combination of the binary elements 0 and 1. If it is known that the arrival of a certain order will take place from 1st until the 16$^{th}$ of March, find the minimal number of the binary elements of each character required for coding this message.

Problem 4: Let us correspond to each letter the number showing its order into the alphabet (A=1, B=2, C=3 etc). Let us correspond also to each word consisting of 4 letters a 2X2 matrix in the obvious way; e.g. the matrix $\begin{bmatrix} 19 & 15 \\ 13 & 5 \end{bmatrix}$ corresponds to the word SOME. Using the matrix E=$\begin{bmatrix} 8 & 5 \\ 11 & 7 \end{bmatrix}$ as an encoding matrix how you could send the message LATE in the form of a camouflaged matrix to a receiver knowing the above process and how he (she) could decode your message?

Problem 5: The demand function $P(Q_d)=25-Q_d^2$ represents the different prices that consumers willing to pay for different quantities $Q_d$ of a good. On the other hand the supply function $P(Q_s)=2Q_s+1$ represents the prices at which different quantities $Q_s$ of the same good will be supplied. If the market's equilibrium occurs at $(Q_0,P_0)$, the producers who would supply at lower price than $P_0$ benefit. Find the total gain to producers'.

Problem 6: A ballot box contains 8 balls numbered from 1 to 8. One makes 3 successive drawings of a lottery, putting back the corresponding ball to the box before the next lottery. Find the probability of getting all the balls that he draws out of the box different.

Problem 7: A box contains 3 white, 4 blue and 6 black balls. If we put out 2 balls, what is the probability of choosing 2 balls of the same colour?

Problem 8: The population of a country is increased proportionally. If the population is doubled in 50 years, in how many years it will be tripled?

Problem 9: A wine producer has a stock of wine greater than 500 and less than 750 kilos. He has calculated that, if he had the double quantity of wine and transferred it to bottles of 12, 25, or 40 kilos, it would be left over 6 kilos each time. Find the quantity of stock.

Problem 10: Among all cylindrical towers having a total surface of $180\pi$ m$^2$, which one has the maximal volume?

Remark: Some students didn't include to the total surface the one base (ground-floor) and they found another solution, while some others didn't include both bases (roof and ground-floor) and they found no solution, since we cannot construct a cylinder with maximal volume from its surrounding surface.